\documentclass[letterpaper, 10 pt, conference]{ieeeconf}  

\IEEEoverridecommandlockouts                              

\usepackage{graphics} 
\usepackage{amsmath} 
\usepackage{amssymb}  
\usepackage{subfiles}
\usepackage{makecell}
\usepackage{graphicx}
\usepackage{multirow}
\usepackage{soul}
\usepackage{soul}
\usepackage{xcolor}
\usepackage[utf8]{inputenc}
\usepackage{hyperref}
\usepackage{cleveref}
\usepackage[T1]{fontenc}
\usepackage{caption}
\usepackage[linesnumbered, ruled]{algorithm2e}
\DontPrintSemicolon
\SetKwComment{Comment}{\footnotesize$\triangleright\,\,$}{}
\SetInd{0.25em}{0.8em}

\SetCommentSty{mycommfont}

\SetFuncSty{myfuncfont}
\usepackage{amssymb,amsmath,caption}
\usepackage[hang,flushmargin]{footmisc}
\usepackage{lipsum}

\crefformat{section}{\S#2#1#3} 
\crefformat{subsection}{\S#2#1#3}
\crefformat{subsubsection}{\S#2#1#3}

\title{\LARGE \bf
\textit{IKLink}: End-Effector Trajectory Tracking with Minimal Reconfigurations
}

\author{Yeping Wang$^{1}$, Carter Sifferman$^{1}$, and Michael Gleicher$^{1}$
\thanks{$^{1}$ All authors are with the Department of Computer Sciences, University of Wisconsin-Madison, Madison, WI 53706, USA $\qquad\qquad\qquad\qquad\qquad\qquad\qquad\qquad\qquad\qquad\qquad\qquad\qquad\qquad\qquad\qquad\qquad\qquad\qquad\qquad\qquad\qquad\qquad\qquad\qquad\qquad\qquad\qquad\qquad\qquad\qquad\qquad\qquad\qquad\qquad\qquad\qquad\qquad\qquad\qquad\qquad\qquad\qquad\qquad\qquad\qquad\qquad\qquad\qquad\qquad\qquad\qquad\qquad\qquad\qquad\qquad\qquad\qquad\qquad\qquad\qquad\qquad\qquad$
{\tt\small [yeping|sifferman|gleicher]@cs.wisc.edu}}%
\thanks{This work was supported by Los Alamos National Laboratory and the Department of Energy.}
}

\makeatletter
\renewcommand\LARGE{\@setfontsize\LARGE{15.6}{19}}
\let\@oldmaketitle\@maketitle
\renewcommand{\@maketitle}{\@oldmaketitle
   \vspace{3mm}
    \includegraphics[width=7in]{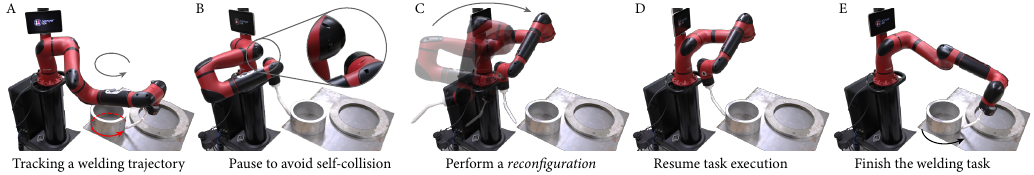}
    \captionof{figure}{We present a method that enables a robot to track provided end-effector trajectories while performing minimal reconfigurations. 
    In this figure, the robot tracks a welding trajectory (red line in A) with a single reconfiguration.
    (B) The robot is unable to proceed due to self-collision avoidance. (C) The robot performs a reconfiguration, where the robot deviates from the task trajectory and moves to a configuration that is far away from self-collision. (D) The robot resumes task execution from the new joint configuration and (E) finishes the task. }
    \label{fig: teaser}
    \vspace{-4mm}
    }

\newcommand{\algorithmfootnote}[2][\footnotesize]{%
  \let\old@algocf@finish\@algocf@finish
  \def\@algocf@finish{\old@algocf@finish
    \leavevmode\rlap{\begin{minipage}{\linewidth}
    #1#2
    \end{minipage}}%
  }%
}
\makeatother

\begin{document}
 \vspace{-10mm}

\maketitle

\addtocounter{figure}{-1} 

\begin{abstract}
Many applications require a robot to accurately track reference end-effector trajectories. Certain trajectories may not be tracked as single, continuous paths due to the robot's kinematic constraints or obstacles elsewhere in the environment. In this situation, it becomes necessary to divide the trajectory into shorter segments. Each such division introduces a \textit{reconfiguration}, in which the robot deviates from the reference trajectory, repositions itself in configuration space, and then resumes task execution. The occurrence of reconfigurations should be minimized because they increase time and energy usage. In this paper, we present \textit{IKLink}, a method for finding joint motions to track reference end-effector trajectories while executing the minimum number of reconfigurations. Our graph-based method generates a diverse set of Inverse Kinematics (IK) solutions for every waypoint on the reference trajectory and utilizes a dynamic programming algorithm to find the optimal motion by linking the IK solutions. We demonstrate the effectiveness of \textit{IKLink} through a simulation experiment and an illustrative demonstration using a physical robot.

\end{abstract}

\vspace{-1mm}
\section{INTRODUCTION}
In applications such as welding, painting, sweeping, or visual tracking,  a robot needs to accurately track a reference trajectory with the end-effector. The problem of finding a joint space motion that drives the end-effector to accurately match a Cartesian space trajectory is called \textit{end-effector trajectory tracking}. While both optimization-based \cite{kang2020torm,holladay2016distance} and graph-based \cite{rakita2019stampede} methods have been presented to solve the problem, most previous approaches operate under the assumption that the end-effector trajectory can be tracked as a single and continuous path, without the need to segment the trajectory into shorter parts. The uninterrupted tracking of some trajectories can be infeasible due to robot kinematics constraints such as self-collision, joint limits, and singularities. 
Furthermore, the assumption is overly strict for some applications, such as welding or inspection, which allow a robot to pause the task execution and resume the same Cartesian pose from another joint configuration. This action, which is referred to as a \textit{reconfiguration} \cite{yang2022optimal}, repositions the robot in its joint space (see Fig. \ref{fig: teaser} for an example). In order to sequentially visit every point on trajectories of varying complexity, a robot must leverage its ability to undertake reconfigurations. However, reconfigurations increase energy usage and time to complete a trajectory, so minimizing their occurrence is of great importance.


In this paper, we introduce a graph-based method that enables a robot to track end-effector trajectories while performing the fewest reconfigurations. The method, called \textit{IKLink}, first generates a diverse set of inverse kinematics (IK) solutions for every waypoint on the reference end-effector trajectory. Subsequently, a dynamic programming algorithm is utilized to find the optimal joint trajectory by linking the IK solutions. We provide an open-source implementation of our proposed method\footnote{\textit{IKLink} is open-sourced: \url{https://github.com/uwgraphics/IKLink}}.

The central contribution of this paper is \textit{IKLink} (\cref{sec:method_overview}, \cref{sec:technical_details}). In addition, we describe two efficient but sub-optimal approaches (\cref{sec:naive_methods}) as baselines. These three approaches are thoroughly evaluated in a simulation experiment (\cref{sec:evaluation}) that involves 4 robots and 70 randomly generated end-effector trajectories. Our results show that \textit{IKLink} generates accurate and smooth motions with fewer reconfigurations when compared to alternative approaches. 
Finally, we conclude this paper with a discussion of the limitations and implications of this work (\cref{sec:discussion}).

\vspace{-1mm}
\section{RELATED WORK}

In this section, we describe related works in the areas of end-effector trajectory tracking and reconfigurations. 

\subsection{End-Effector Trajectory Tracking} \label{sec:tracking_literature}

The problem of finding a joint space trajectory that drives the end-effector to accurately match a Cartesian space trajectory is known as path-wise inverse kinematics \cite{rakita2019stampede}, task-space non-revisiting tracking \cite{yang2022optimal}, or task-constrained motion planning \cite{cefalo2013task}. Various methods have been proposed in the literature to address the problem, including greedy methods (local optimization), global trajectory optimization, and graph-based approaches. 

The end-effector trajectory tracking problem is conventionally addressed using an Inverse Kinematics (IK) solver that iteratively finds \textit{locally} optimal solutions. These greedy approaches can achieve local objectives such as joint velocities minimization (\textit{e.g.}, via Moore-Penrose pseudoinverse of the Jacobian matrix \cite{siciliano1990kinematic}) or singularity
and self-collision avoidance (via non-linear optimization approaches \cite{praveena2019user}).
While being computationally efficient, these approaches can be myopic and get stuck in local minima \cite{yoon2023learning, holladay2019minimizing, cefalo2013planning}.

The end-effector trajectory tracking problem can be cast as a trajectory optimization problem which directly optimizes a trajectory in joint space while satisfying constraints. Holladay \textit{et al.} \cite{holladay2016distance} utilize a trajectory optimization method, TrajOpt \cite{schulman2014motion}, to minimize the Fréchet distances between current solutions and a given reference trajectory. In order to improve precision, TORM \cite{kang2020torm} minimizes the summed Euclidean distance between every pair of waypoints on the current and the reference trajectories. These trajectory optimization methods are sensitive to initial trajectory \cite{yoon2023learning} and prone to be stuck in local minima \cite{holladay2016distance}. In addition, these approaches are designed to generate a single, continuous solution, \textit{e.g.}, TORM \cite{kang2020torm} has a smoothness objective and TrajOpt \cite{schulman2014motion} has a displacement minimization objective to ensure joint space continuity. Therefore, trajectory optimization methods are designed to track a trajectory without interruptions and fail when tracking a long or complex end-effector trajectory that requires reconfigurations.

Alternatively, graph-based approaches construct a hierarchical graph where each layer consists of a set of IK solutions for each waypoint on the end-effector trajectory. These methods can be classified into two groups according to the type of graph utilized. The first group of methods connects an IK solution to \textit{multiple} IK solutions in the previous layer that satisfy the velocity constraints, resulting in a Directed Acyclic Graph (DAG). Prior works employ various shortest path algorithms to find the optimal joint trajectory in the DAG, including value iteration dynamic programming \cite{rakita2019stampede}, Dijkstra’s algorithm \cite{holladay2019minimizing, alatartsev2014improving, Descartes}, and
Lifelong Planning A* \cite{niyaz2020following}. The second group of approaches builds \textit{trees} by connecting an IK solution to the \textit{nearest} IK solution in the previous layer
\cite{cefalo2013task, cefalo2013planning, oriolo2002probabilistic}. A joint trajectory is obtained by traversing from a leaf to the root. 
Graph-based methods are generally computationally expensive \cite{holladay2019minimizing, malhan2022generation} and various clustering methods have been used to reduce the number of vertices in the graph \cite{rakita2019stampede, malhan2022generation}. In contrast to the above work that tracks a given end-effector trajectory \textit{uninterruptedly}, this paper presents a graph-based approach that constructs trees and uses dynamic programming to track an end-effector trajectory while performing minimal reconfigurations.

\subsection{Reconfigurations}

The aforementioned end-effector trajectory tracking methods assume that the given end-effector trajectory can be tracked as a single, continuous path. This assumption is invalid in cases where a continuous motion does not exist in the joint space to track the given end-effector trajectory. Additionally, the assumption is overly restrictive on tasks that allow \textit{reconfigurations}.
Prior works propose methods to generate robot motions that include minimal reconfigurations in end-effector trajectory tracking \cite{yang2022optimal} and area coverage \cite{yang2020cellular, yang2020non}. However, these methods are designed for non-redundant robots and it is unclear how to apply them to redundant robots. The method proposed in this paper is applicable to both non-redundant and redundant robots.

\section{TECHNICAL OVERVIEW} \label{sec:technical_overview}
In this section, we formalize our problem statement, describe two baseline approaches, and provide an overview of our proposed method.
\subsection{Problem Formulation} 
Consider an $k$ degree of freedom robot whose joint configuration and end-effector pose are denoted by $\mathbf{q} \in \mathbb{R}^k$ and $\mathbf{p} \in SE(3)$, respectively. The goal of this work is to compute a joint space trajectory $\xi: [0,T_\chi] \rightarrow \mathbb{R}^k$ that drives the robot to match a desired end-effector trajectory $\chi: [0,T_\chi] \rightarrow SE(3)$ and has a minimum number of joint-space discontinuities. A joint space discontinuity occurs between time $t$ and $t + \tau$ if there exists one joint $j$ whose velocity $[\xi(t+\tau)_j - \xi(t)_j]/\tau$ has exceeded its velocity limit, where $\tau > 0$ is a short time interval. In practice, the target end-effector trajectory $\chi$ is generally represented in a discrete form $\{(t_0, \mathbf{p}_0), (t_1, \mathbf{p}_1), ..., (t_{n-1}, \mathbf{p}_{n-1})\}$, where $\mathbf{p}_i$ is the end-effector pose at timestamp $t_i$. We assume that the sampling is dense enough that the $n$ waypoints can accurately approximate the end-effector trajectory. In addition, we assume that the robot's task permits reconfiguration and we can find motions to execute reconfigurations.

\subsection{Baseline Approaches} \label{sec:naive_methods}

\begin{figure*} [t]
  \centering
  \includegraphics[width=7in]{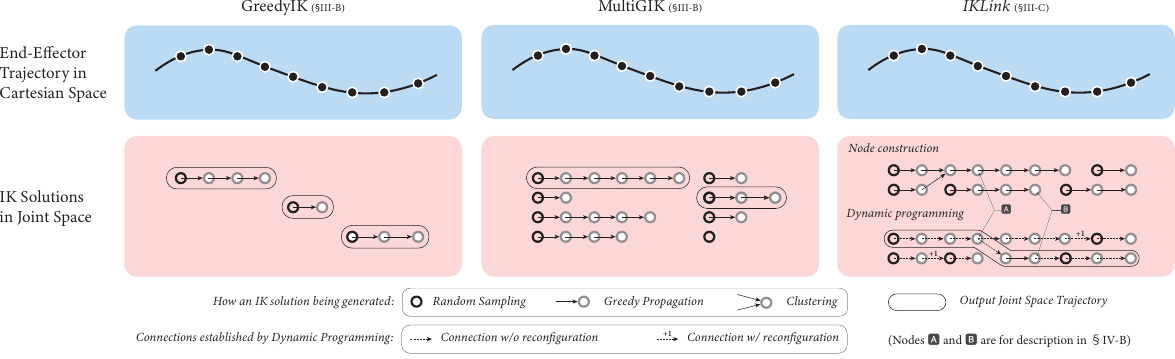}
  \caption{An illustration of the three approaches described in this paper and evaluated in our experiment.}
  \label{fig: approaches}
  \vspace{-5mm}
\end{figure*}

In order to facilitate understanding of the problem, we describe two baseline approaches. These two approaches serve as baselines in our experiment in \cref{sec:evaluation}. Fig. \ref{fig: approaches} provides an illustration of the two baseline approaches as well as the method proposed in this paper. 

\textit{GreedyIK} --- Perhaps the most straightforward solution is to use an Inverse Kinematics (IK) solver to keep the robot on the reference end-effector trajectory, optimizing for the closest solution from the previous state. In the event that the IK solver is unable to proceed due to kinematic constraints, the robot reconfigures by reinitializing the IK solver from a \textit{randomly} sampled configuration. While being simple and computationally efficient, GreedyIK has three primary drawbacks: 1) it is myopic because it only finds \textit{locally} optimal solutions; 2) due to the nonlinearity of an IK problem, the IK solver may fail to find a valid IK solution; 3) the strategy of undergoing reconfiguration to a random IK solution is clearly not optimal. In practice, GreedyIK generally performs reconfigurations more frequently than necessary.

\textit{MultiGIK} --- This approach involves the instantiation of multiple greedy IK solvers in parallel and keeps the \textit{longest} solution, \textit{i.e.}, the solution that can track the given end-effector trajectory for the maximum duration. This procedure is repeated starting from the subsequent waypoint of the longest solution.
MultiGIK has good, but not optimal, performance. Yang \textit{et al.} \cite{yang2022optimal} have proven that choosing longest possible motion segments leads to a solution with the minimum number of reconfigurations. 
However, the nested greedy IK solver in MultiGIK is myopic and not guaranteed to find the longest possible motion, so MultiGIK is not optimal. MultiGIK finds optimal solutions only when its greedy motion coincides with the longest possible motion. In practice, MultiGIK works well, so we use it as a baseline in our experiment in \cref{sec:evaluation}. Our results show that IKLink consistently finds motions with equal or fewer reconfigurations than MultiGIK. 

\subsection{Overview of IKLink} \label{sec:method_overview}

Our algorithm consists of two stages. First, it generates a number of inverse kinematics solution candidates for each waypoint on the target end-effector trajectory. The output of this stage is an $n \times m$ IK solution table $S$, where $n$ is the number of waypoints on the reference end-effector trajectory and $m$ is the number of IK candidates for each waypoint. 
Afterward, a dynamic programming algorithm is utilized to establish connections between solution candidates and compute the solution with minimal reconfigurations. We name the method \textit{IKLink} because it generates joint trajectories by linking IK solutions. The efficacy of the dynamic programming algorithm depends upon the diversity of the IK solutions. We will describe our sampling strategy and the dynamic programming algorithm in the next section.

\section{TECHNICAL DETAILS} \label{sec:technical_details}
In \cref{sec:method_overview}, we gave an abstract overview of \textit{IKLink}. We elaborate on the details of our method in this section. 

\subsection{IK Solution Construction} \label{sec:IK_construction}
\textit{IKLink} generates robot motions by linking Inverse Kinematics (IK) solutions, so the quality of the IK solutions directly impacts the algorithm's performance. Our goal is to sample a diverse set of IK solutions because similar IK solutions can result in unnecessary computation and hinder performance of the dynamic programming algorithm.
Inspired by Stampede \cite{rakita2019stampede}, we use three methods to construct solution candidates: random sampling, greedy propagation using an optimization-based IK solver, and clustering.  

\textit{Random Sampling} -- The IK solutions of the first waypoint on the target end-effector trajectory are randomly sampled. 
We generate the samples using Trac-IK \cite{beeson2015trac} by uniformly sampling the seeds from the robot's joint space. 

\textit{Greedy Propagation} -- To facilitate smooth motions, we utilize greedy propagation instead of random sampling to generate solution candidates for the subsequent waypoints on the trajectory. Greedy propagation uses a greedy, optimization-based IK solver to find the closest IK solution from a preceding solution candidate.
We choose RelaxedIK \cite{rakita2018relaxedik} as the greedy IK solver for its ability to generate smooth motions. RelaxedIK achieves higher accuracy the more iterations it is allowed. In our prototype, to balance accuracy and computing time, we consider an IK solution to be legitimate if the positional and rotational error of the end-effector are smaller than 1 mm and 0.01 rad, respectively.

\begin{figure*} [tb]
  \centering
  \includegraphics[width=7in]{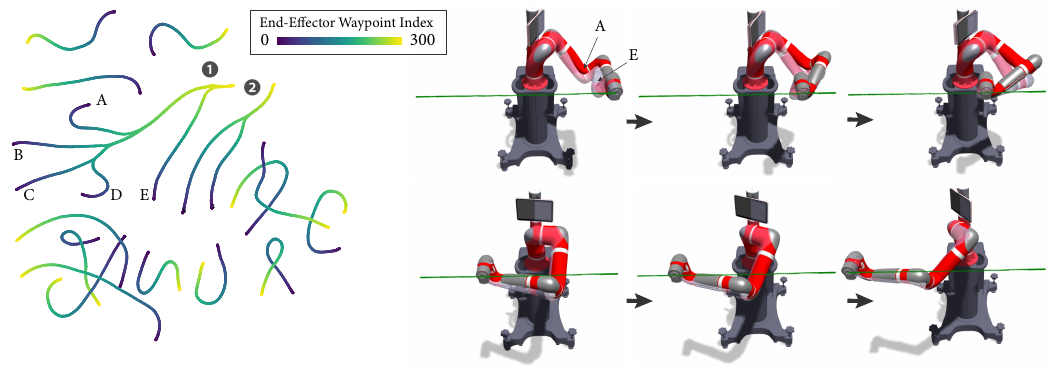}
  \vspace{-7mm}
  \caption{Left: A UMAP visualization of 20 robot motions that track an identical end-effector trajectory. Each motion starts with a random configuration and greedily propagates to the subsequent end-effector pose using an optimization-based IK solver. 
  Convergence of motions occurs in regions 1 and 2. Right: We visualize motions A and E using a solid color robot and a translucent robot, respectively. The two motions start from different configurations but converge together over the course of the trajectory.
  This figure shows that greedy propagation does not consistently yield diverse solutions.}
  \label{fig: ik_converage}
  \vspace{-5mm}
\end{figure*}

However, the solution candidates created by greedy propagation have a tendency to converge over the course of the trajectory. To illustrate this, we command a Rethink Robotics Sawyer robot to track a straight line, starting from 20 different joint configurations. This results in 20 different motions using greedy propagation. Fig. \ref{fig: ik_converage} visualizes these motions using the Uniform Manifold Approximation and Projection (UMAP) method \cite{mcinnes2018umap} for dimensionality reduction. The observation implies that the IK solutions generated by greedy propagation have a tendency to converge, hence not consistently yielding a diverse set of solutions.

\textit{Clustering} -- To avoid repeated candidates and maintain solution diversity, we merge similar IK solutions before greedily propagating to the IK solutions for the next waypoint. We utilize Density-Based Spatial Clustering of Applications with Noise (DBSCAN) \cite{ester1996density} to merge IK solutions that are closer than 0.05 rad in joint space. Following the merging process, the newly freed spots in the IK solution table are filled by randomly sampled solutions.

\subsection{Dynamic Programming}

In the preceding subsection, we described the procedure of constructing IK solutions for each waypoint along the target end-effector trajectory. Below, we present a dynamic programming approach to establish connections between IK solutions, ultimately resulting in a motion.

\begin{algorithm}
    \caption{Dynamic Programming}
    \label{alg:ad_hoc}
    \SetKwProg{Fn}{Function}{}{}
    \SetKwProg{For}{for}{\ do}{}
    \SetKwProg{If}{if}{\ then}{}
    \SetKwInOut{Input}{input}
    \SetKwInOut{Output}{output}
    \SetKw{KwTo}{to}
    \SetKw{KwOr}{or}
    \SetKw{KwAnd}{and}
    \SetKwFunction{FDP}{DP}%
    \SetKwFunction{FCC}{CHECK\_CONT}%
    \SetKwFunction{FJD}{JNT\_DIST}%
    \Input{an $n \times m$ table $S$ that contains IK solutions \\ an $n$ array $t$ that contains timestamps}
    \Output{a joint space trajectory that has the minimum number of discontinuities }
        $c[0..n, 0..m] = 0$ \Comment*[r]{num of reconfig}
        $p[0..n, 0..m] = 0$ \Comment*[r]{predecessor}
        $l[0..n, 0..m] = 0$ \Comment*[r]{length in joint space}
        \For ( \Comment*[f]{start linking IK solutions}) {$x = 1$ \KwTo $n{-}1$} { 
             $\Delta t \gets t[x] {-} t[x{-}1]$ \;
            \For {$y_1 = 0$ \KwTo $m{-}1$} {
                $c[x,y_1],l[x,y_1] \gets \infty, \infty$ \; 
                \For {$y_2 = 0$ \KwTo $m{-}1$} {
                    \uIf {$c[x{-}1,y_2]{+}1{<}c[x,y_1]$ 
                        \KwOr $(c[x{-}1, y_2]{+}1$ ${==}c[x,y_1]$  
                        \KwAnd  $l[x{-}1,y_2]{<}l[x,y_1])$}  {
                        \Comment*[r]{link w/ reconfiguration}
                        $c[x,y_1] \gets c[x{-}1, y_2]{+}1$ \;
                        $p[x,y_1],l[x,y_1] \gets y_2,l[x{-}1,y_2]$ \;
                    }
                    \uIf(\Comment*[f]{link w/o reconfiguration}){\FCC{$S[x{-}1,y_2],S[x,y_1]$, $\Delta t$} } { 
                        $d \gets l[x{-}1,y_2]{+}\lVert S[x{-}1,y_2] - S[x,y_1] \rVert_2$\;
                        \uIf {$c[x{-}1,y_2]{<}c[x,y_1]$ \KwOr $(c[x{-}1,y_2]{==}$ $c[x,y_1]$ \KwAnd $\,d{<}l[x,y_1])$}{
                            $c[x,y_1] \gets c[x{-}1,y_2]$ \;
                            $p[x,y_1],l[x,y_1] \gets y_2,d$ 
                        }
                    }
                } 
            }
        }
        $c_{min}, l_{min}, y_{min} \gets \infty, \infty, \infty$\;
        \For(\Comment*[f]{find index of optimal motion}) {$y = 0$ \KwTo $m{-}1$ }{  
                \uIf  {$c[n{-}1, y] {<} c_{min}$ \KwOr $(c[n{-}1, y]{==} c_{min}$ \KwAnd $\,l[n{-}1,y]{<}l_{min})$}{
                $c_{min},l_{min}, y_{min} \gets c[n{-}1, y], l[n{-}1, y], y$\;
                }
            }
        traj $\gets [\,]$  \Comment*[r]{optimal motion}
        \For {$x = n{-}1$ \KwTo $0$}{
            Append ($t[x]$, $S[x, y_{min}]$) to traj \;
            $y_{min} \gets p[x, y_{min}]$ \;
        }
        \textbf{return} Reverse(traj)


    \algorithmfootnote{Function {\footnotesize CHECK\_CONT} returns true if the robot can move between the two given joint configurations within the time duration, without exceeding joint velocity limits.}
\end{algorithm}

The dynamic programming method takes the IK solution table $S$ as input. For each cell in the table, we use $c[x,y]$ to denote the minimum number of reconfigurations needed to travel from the first column of the table (the start of the reference end-effector trajectory) to the current IK solution $S[x,y]$. The predecessor of $S[x,y]$ is stored in $p[x,y]$ and the length in joint space from the first column to $S[x,y]$ is stored in $l[x,y]$. As described in Algorithm 1, the dynamic programming method proceeds by computing $c,p$, and $l$ from the first column of the table. After traversing through all the columns, the optimal joint trajectory can be traced backward using the stored predecessors $p$. 

The efficacy of the dynamic programming algorithm stems from the fact that the problem can be broken down into sub-problems of finding an optimal solution that ends with a specific joint configuration for a sub-trajectory $\chi_{[0:T]}$, where $T<T_\chi$.  Let us assume $\xi^*$ is an optimal solution for the entire end-effector trajectory $\chi$, then $\xi^*_{[0:T]}$ must be an optimal solution for $\chi_{[0,T]}$ among all solutions ended with $\xi^*(T)$. If there exists a solution superior to $\xi^*_{[0:T]}$ and it also ends with $\xi^*(T)$, a superior global solution can be obtained by substituting the new solution in $\xi^*$ for $\xi^*_{[0:T]}$. This derives a contradiction from the assumption that $\xi^*$ is optimal.

The objective of the dynamic programming algorithm is to \textit{link} IK solutions. Its superiority becomes evident when greedy propagation becomes myopic or fails to find a valid IK solution. Fig. \ref{fig: approaches} provides an example for each situation: during node construction, Node A myopically propagates to the next node in the same row; later, the dynamic programming algorithm links it to a node in the next row that leads to fewer reconfigurations. In another example, during node construction, greedy propagation fails to find a successive IK solution for Node B; later, the dynamic programming algorithm finds a randomly sampled node as a successor node, which avoids the need to do a reconfiguration.

\vspace{-1mm}
\section{Evaluation} \label{sec:evaluation}
In this section, we compare our method, \textit{IKLink}, with two alternative approaches described in \cref{sec:naive_methods}, GreedyIK and MutliGIK, on four benchmark tasks. In our prototype systems, MutliGIK instantiates 300 IK solvers in parallel and \textit{IKLink} samples 300 IK solutions for each waypoint on the reference trajectory ($m=300$). 
To more fairly assess \textit{IKLink} over the greedy methods, we provide an additional baseline that takes a similar amount of time to \textit{IKLink}.
The MultiGIK$\times$30 baseline uses the MultiGIK method but instantiates 300$\times$30 IK solvers in parallel (rather than 300).

\subsection{Implementation Details}

Our prototype implementation is based on the open-source RangedIK\footnote{IK solver: \url{https://github.com/uwgraphics/relaxed\_ik\_core/tree/ranged-ik}} library. RangedIK \cite{wang2023rangedik} is an extension of RelaxedIK \cite{rakita2018relaxedik} which can effectively exploit tolerances in joint or Cartesian space. 
We use RangedIK without tolerances being specified, making it functionally equivalent to RelaxedIK. 
To ensure fair comparisons, all three approaches used the same function for IK solution sampling and greedy propagation.  All evaluations were conducted on a laptop with Intel Core i7-11800H 2.30 GHz CPU and 16 GB of RAM.

\begin{figure*} [tb]
  \centering
  \includegraphics[width=7in]{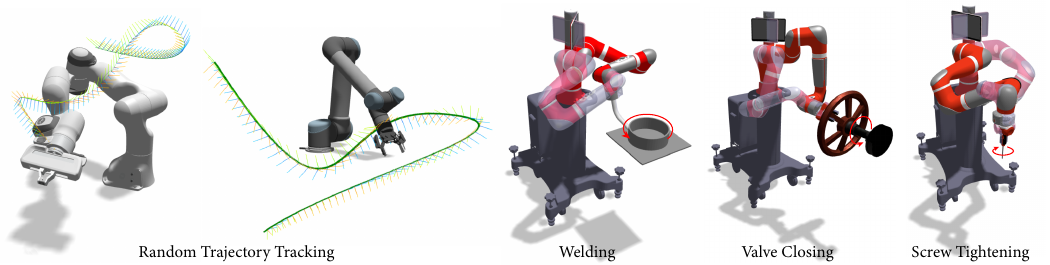}
  \vspace{-5mm}
  \caption{Our experiment involves four benchmark applications. In Random Trajectory Tracking, the reference end-effector trajectories are visualized in green curves, with coordinate frames attached to show orientations. In the other three benchmarks, the translucent robot and the solid-color robot show the configurations before and after a reconfiguration. These visualizations were generated using \textit{Motion Comparator}\protect\footnotemark.}
  \label{fig: tasks}
  \vspace{-1mm}
\end{figure*}

\begin{table*}[tb]
\caption{Experiment Results and Metrics of the Input Trajectories} 
\vspace{-3mm}
\label{tab:results}
\begin{center}
\begin{tabular}{c|c|l|rr|lll|ccc}
\hline
\multicolumn{2}{l|}{\multirow{3}{*}{Benchmark}} & \makecell[c]{\multirow{3}{*}{Method}} & \makecell[c]{\multirow{3}{*}{\makecell[l]{Mean Num\\ of Reconfig}}} & \multirow{3}{*}{\makecell{Mean Comput-\\ation Time (s)$^\dagger$}} & \multicolumn{3}{c|}{Output Joint Motion Metrics} & \multicolumn{3}{c}{Input End-Effector Trajectory Metrics} \\
\multicolumn{2}{l|}{} & &  &  & \makecell{Mean Joint \\ Vel. (rad/s) } &  \makecell{Max Pos. \\ Error (m)$^\ddagger$} & \makecell{Max Rot. \\ Error (rad)$^\ddagger$} & \makecell{Mean \# of \\ Waypoints} & \makecell{Mean\\Len. (m)}&  \makecell{Mean\\Rot. (rad)}\\ 
\hline

\rule{0pt}{1.1\normalbaselineskip}%

\rule{0pt}{1.1\normalbaselineskip}%
\multirow{16}{*}{\rotatebox[origin=c]{90}{Random Trajectory Tracking}}%
&\multirow{4}{*}{\rotatebox[origin=c]{90}{iiwa}}%
&GreedyIK& 13.80$\pm$6.87 & 0.13$\pm$0.03 & 0.17$\pm$0.04 & 1e-3$\pm$6e-6 & 1e-2$\pm$9e-6 & \multirow{4}{*}{\makecell[l]{685.6\\$\pm$128.2}} & \multirow{4}{*}{\makecell[l]{2.28\\$\pm$0.43}} & \multirow{4}{*}{\makecell[l]{9.91\\$\pm$2.50}} \\   
&&MultiGIK& 1.90$\pm$0.94 & 6.34$\pm$2.75 & 0.18$\pm$0.04 & 1e-3$\pm$1e-6 & 1e-2$\pm$5e-6 & & & \\  
&&$^\lfloor$\scriptsize MultiGIK$\times$30& 1.60$\pm$1.02 & 186.01$\pm$70.0 & 0.18$\pm$0.04 & 1e-3$\pm$3e-6 & 1e-2$\pm$1e-5 & & & \\  
&&\textit{IKLink}& \textbf{0.40$\pm$0.80} & 166.55$\pm$140 & 0.18$\pm$0.04 & 1e-3$\pm$3e-5 & 1e-2$\pm$5e-6 & & & \\   \cline{2-11} 

\rule{0pt}{1.1\normalbaselineskip}%
&\multirow{4}{*}{\rotatebox[origin=c]{90}{Sawyer}}%
&GreedyIK& 14.20$\pm$11.6 & 0.12$\pm$0.03 & 0.22$\pm$0.04 & 1e-3$\pm$2e-5 & 1e-2$\pm$1e-5 & \multirow{4}{*}{\makecell[l]{787.5\\$\pm$169.6}} & \multirow{4}{*}{\makecell[l]{2.62\\$\pm$0.57}} & \multirow{4}{*}{\makecell[l]{11.60\\$\pm$2.97}} \\   
&&MultiGIK& 1.50$\pm$0.67 & 7.05$\pm$1.71 & 0.21$\pm$0.06 & 1e-3$\pm$1e-5 & 1e-2$\pm$2e-5 & & & \\  
&&$^\lfloor$\scriptsize MultiGIK$\times$30& 1.70$\pm$1.85 & 212.68$\pm$72.5 & 0.21$\pm$0.04 & 1e-3$\pm$6e-6 & 1e-2$\pm$3e-5 & & & \\  
&&\textit{IKLink}& \textbf{1.10$\pm$0.30} & 119.25$\pm$33.2 & 0.19$\pm$0.06 & 1e-3$\pm$1e-5 & 1e-2$\pm$3e-5 & & & \\ \cline{2-11} 

\rule{0pt}{1.1\normalbaselineskip}%
&\multirow{4}{*}{\rotatebox[origin=c]{90}{Panda}}%
&GreedyIK& 16.60$\pm$8.24 & 0.14$\pm$0.04 & 0.23$\pm$0.04 & 1e-3$\pm$5e-6 & 1e-2$\pm$1e-5 & \multirow{4}{*}{\makecell[l]{669.3\\$\pm$107.2}} & \multirow{4}{*}{\makecell[l]{2.23\\$\pm$0.36}} & \multirow{4}{*}{\makecell[l]{11.95\\$\pm$2.66}} \\   
&&MultiGIK& 4.40$\pm$1.50 & 10.81$\pm$5.11 & 0.23$\pm$0.03 & 1e-3$\pm$3e-5 & 1e-2$\pm$1e-5 & & & \\   
&&$^\lfloor$\scriptsize MultiGIK$\times$30& 3.70$\pm$1.55 & 376.52$\pm$150 & 0.24$\pm$0.03 & 1e-3$\pm$5e-5 & 1e-2$\pm$1e-5 & & & \\   
&&\textit{IKLink}& \textbf{1.70$\pm$0.90} & 219.92$\pm$61.1 & 0.24$\pm$0.04 & 1e-3$\pm$7e-6 & 1e-2$\pm$3e-5 & & & \\   
\cline{2-11}

\rule{0pt}{1.1\normalbaselineskip}%
&\multirow{4}{*}{\rotatebox[origin=c]{90}{UR5}}%
&GreedyIK& 16.00$\pm$14.1 & 0.11$\pm$0.03 & 0.26$\pm$0.03 & 1e-3$\pm$9e-6 & 1e-2$\pm$2e-5 & \multirow{4}{*}{\makecell[l]{713.0\\$\pm$148.2}} & \multirow{4}{*}{\makecell[l]{2.37\\$\pm$0.49}} & \multirow{4}{*}{\makecell[l]{11.77\\$\pm$1.89}} \\   
&&MultiGIK& 2.10$\pm$1.30 & 6.62$\pm$3.57 & 0.27$\pm$0.03 & 1e-3$\pm$3e-5 & 1e-2$\pm$2e-5 & & & \\   
&&$^\lfloor$\scriptsize MultiGIK$\times$30& 1.60$\pm$1.20 & 162.88$\pm$61.6 & 0.27$\pm$0.04 & 1e-3$\pm$4e-5 & 1e-2$\pm$2e-5 & & & \\   
&&\textit{IKLink}& \textbf{1.50$\pm$1.12} & 121.87$\pm$29.0 & 0.25$\pm$0.03 & 1e-3$\pm$6e-5 & 1e-2$\pm$8e-6 & & & \\   \hline

\rule{0pt}{1.1\normalbaselineskip}%
\multirow{4}{*}{\rotatebox[origin=c]{90}{Weld}}%
&\multirow{12}{*}{\rotatebox[origin=c]{90}{Sawyer}}%
&GreedyIK& 28.50$\pm$15.4 & 0.13$\pm$0.02 & 0.16$\pm$0.02 & 1e-3$\pm$5e-5 & 1e-2$\pm$3e-6 & \multirow{4}{*}{\makecell[l]{450.0\\$\pm$0.0}} & \multirow{4}{*}{\makecell[l]{0.86\\$\pm$0.17}} & \multirow{4}{*}{\makecell[l]{6.27\\$\pm$0.00}} \\   
& &MultiGIK& 1.50$\pm$1.28 & 5.30$\pm$1.13 & 0.16$\pm$0.02 & 1e-3$\pm$3e-5 & 1e-2$\pm$2e-6 & & & \\   
& &$^\lfloor$\scriptsize MultiGIK$\times$30& 1.20$\pm$0.98 & 142.12$\pm$31.4 & 0.16$\pm$0.02 & 1e-3$\pm$2e-5 & 1e-2$\pm$1e-5 & & & \\   
& &\textit{IKLink}& \textbf{0.90$\pm$0.30} & 88.98$\pm$11.3 & 0.14$\pm$0.03 & 9e-4$\pm$8e-5 & 1e-2$\pm$3e-6 & & & \\   \cline{0-0} \cline{3-11} 

\rule{0pt}{1.1\normalbaselineskip}%
\multirow{4}{*}{\rotatebox[origin=c]{90}{Valve}}%
&&GreedyIK& 22.60$\pm$9.17 & 0.16$\pm$0.04 & 0.35$\pm$0.02 & 1e-3$\pm$7e-6 & 1e-2$\pm$2e-6 & \multirow{4}{*}{\makecell[l]{585.0\\$\pm$105.0}} & \multirow{4}{*}{\makecell[l]{3.67\\$\pm$0.66}} & \multirow{4}{*}{\makecell[l]{24.46\\$\pm$4.40}} \\   
&&MultiGIK& 2.50$\pm$0.81 & 7.77$\pm$1.68 & 0.36$\pm$0.03 & 1e-3$\pm$3e-5 & 1e-2$\pm$6e-6 & & & \\ 
&&$^\lfloor$\scriptsize MultiGIK$\times$30& \textbf{2.30$\pm$0.90} & 227.53$\pm$57.4 & 0.36$\pm$0.02 & 1e-3$\pm$2e-5 & 1e-2$\pm$5e-6 & & & \\    
&&\textit{IKLink}& \textbf{2.30$\pm$0.90} & 110.38$\pm$22.7 & 0.35$\pm$0.02 & 1e-3$\pm$4e-5 & 1e-2$\pm$7e-6 & & & \\   \cline{0-0} \cline{3-11} 

\rule{0pt}{1.1\normalbaselineskip}%
\multirow{4}{*}{\rotatebox[origin=c]{90}{Screw}}%
&&GreedyIK& 22.10$\pm$11.8 & 0.28$\pm$0.11 & 0.25$\pm$0.01 & 1e-3$\pm$4e-5 & 1e-2$\pm$6e-6 & \multirow{4}{*}{\makecell[l]{1155.0\\$\pm$285.0}} & \multirow{4}{*}{\makecell[l]{0.03\\$\pm$0.01}} & \multirow{4}{*}{\makecell[l]{48.34\\$\pm$11.94}} \\   
&&MultiGIK& 4.70$\pm$1.35 & 21.58$\pm$5.94 & 0.24$\pm$0.01 & 1e-3$\pm$3e-5 & 1e-2$\pm$8e-6 & & & \\ 
&&$^\lfloor$\scriptsize MultiGIK$\times$30& \textbf{4.60$\pm$1.20} & 647.97$\pm$176 & 0.25$\pm$0.01 & 1e-3$\pm$8e-6 & 1e-2$\pm$1e-5 & & & \\    
&&\textit{IKLink}& \textbf{4.60$\pm$1.20} & 230.59$\pm$80.3 & 0.26$\pm$0.01 & 1e-3$\pm$1e-5 & 1e-2$\pm$2e-5 & & & \\   
   \hline 


\hline
\multicolumn{11}{l}{\rule{0pt}{1\normalbaselineskip}%
The range values are standard deviations. } \\
\multicolumn{11}{l}{
$\dagger$: All methods were implemented in Python and we expect reduced computation time using a more efficient compiled language such as C++. } \\
\multicolumn{11}{l}{\rule{0pt}{1\normalbaselineskip}%
\makecell[l]{$\ddagger$: In our prototype system, we set the positional and rotational tolerance of IK solvers to be 1e-3 m and 1e-2 rad, respectively. As discussed in\\ \cref{sec:IK_construction}, the accuracy can be improved by allowing more computing time for greedy propagation. }}
\vspace{-7mm}
\end{tabular}
\end{center}
\end{table*}

\subsection{Benchmark}
We developed four benchmark applications to compare our method against alternative approaches.

\subsubsection{Random Trajectory Tracking} 
The robot tracks an end-effector trajectory that consists of two consecutive cubic Bézier curves in $SE(3)$, whose control points are randomly sampled within a robot's workspace. We use the cumulative Bézier quaternion curve \cite{kim1995general} to generate a smooth, time-continuous curve in rotation space. The end-effector trajectory is discretized to 300 waypoints per meter. To ensure that every waypoint on the trajectory is reachable, we use Trac-IK \cite{beeson2015trac} for reachability check.  

\subsubsection{Welding} 
The robot welds a vertical cylinder on a horizontal plane. The radius of the cylinder is uniformly sampled from 0.1m to 0.2m and its position is randomly sampled in the workspace in front of the robot.

\subsubsection{Screw Tightening}
The robot uses a screwdriver to rotate a vertically oriented screw in a clockwise direction. The length of the screw is uniformly sampled from 0.02m to 0.04m and the number of turns to tighten the screw is uniformly sampled from 5 to 10. 

\subsubsection{Valve Closing} 
The robot rotates a valve clockwise. The radius of the handle is 0.15m and the number of turns is uniformly sampled from 3 to 5.

\subsection{Experimental Procedure}

We repeat the random trajectory tracking benchmark 10 times on three 7-DoF redundant robots (Rethink Robotics Sawyer, Franka Panda, and KUKA LBR iiwa) and a 6-DoF non-redundant robot (Universal Robot UR5). The welding, screw, and valve benchmarks are repeated 10 times on a Sawyer robot. A total of 70 end-effector trajectories are generated and their waypoint counts, lengths, and angular displacements are reported in Table \ref{tab:results}. We use various approaches to track these trajectories and measure the generated motions. In addition to the number of configurations, we report mean computation time and evaluate motion qualities using mean joint velocity (rad/s), maximum position error (m), and maximum rotation error (rad). 

\subsection{Results}
\begin{figure} [h!]
  \centering
  \includegraphics[width=3.4in]{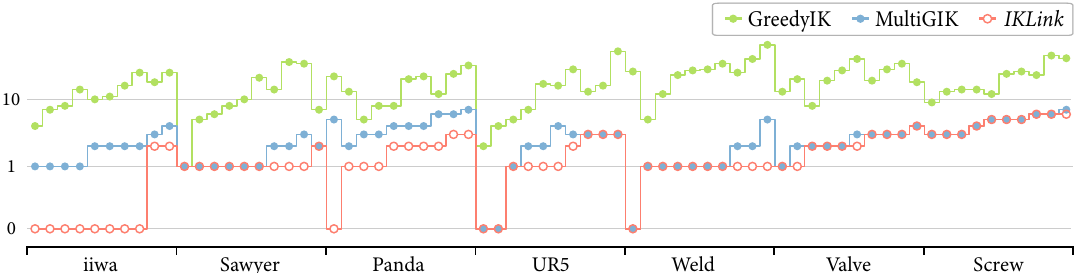}
  \vspace{-5mm}
  \caption{In our experiment, \textit{IKLink} \textit{consistently} generates motions with equal or fewer reconfigurations compared to the alternative approaches. The $x$-axis of the plot corresponds to a total of 70 end-effector trajectories.
  The \textit{y}-axis shows the number of reconfigurations in logarithmic scale.  }
  \label{fig: radar}
  \vspace{-3mm}
\end{figure}

As shown in Fig. \ref{fig: radar}, among all 70 benchmark end-effector trajectories, solutions found by \textit{IKLink} consistently require equal or fewer reconfigurations than those found by GreedyIK and MultiGIK. The results demonstrate the efficacy of \textit{IKLink} at generating motions with minimal reconfigurations. 
Table \ref{tab:results} also shows that \textit{IKLink}'s performance is equal or superior to MultiGIK$\times$30, despite the latter utilizing a larger time budget. In addition, all these approaches generate accurate and smooth motions. 
While speed may not be the most important criterion for offline algorithms, \textit{IKLink} is slower than GreedyIK and MultiGIK. Our implementation has not been optimized for execution speed. 

\subsection{Real-Robot Demonstration}
To further demonstrate the effectiveness of \textit{IKLink}, we generate 
a welding motion using \textit{IKLink} and execute it on a physical Rethink Robotics Sawyer robot. The robot is able to accurately and smoothly track the welding trajectory with one reconfiguration.
We use RRT$^*$ \cite{karaman2011sampling} to plan the motion for the reconfiguration. The demonstration is shown in Fig. \ref{fig: teaser} and the supplementary video\footnote{\url{https://youtu.be/EB4bJ6rJtnY}}.


\section{Discussion} \label{sec:discussion}
In this paper, we present a graph-based method for finding joint motions to track reference end-effector trajectories while undertaking minimal
reconfigurations. Below, we discuss the limitations and implications of this work.

\subsection{Limitations}
Our work has several limitations that highlight potential future directions. First, the goal of this paper is to find a motion with the minimum \textit{number} of reconfigurations. One underlying assumption is that all reconfigurations are equally detrimental, which may not always hold true. The costs to perform reconfigurations may vary depending on the time and energy usage. When the robot is in a cluttered environment, certain reconfigurations may even be infeasible to perform. Therefore, future work should explore methods that integrate the planning of both trajectory tracking motions and reconfiguration motions. Second, the motion qualities of \textit{IKLink} build upon the diversity of Inverse Kinematics (IK) samplers and \textit{IKLink}'s time complexity is $\mathcal{O}(m^2n)$, where $m$ is the number of IK samplers per waypoint. Hence, both \textit{IKLink}'s motion qualities and performance can be improved by using a diverse and efficient IK sampling method, \textit{e.g.}, IKFlow \cite{ames2022ikflow}. Third, our method involves sampling IK solutions within a robot's joint limits and can not sample solutions for revolute joints without position limits. Future work can extend our work to exploit joints with unlimited range. 


\negmedspace
\negmedspace
\negmedspace
\footnotetext{Visualization tool: \url{https://github.com/uwgraphics/MotionComparator}}

\subsection{Implications}

\textit{IKLink} enables a robot to track end-effector trajectories of any complexity while performing minimal reconfigurations. \textit{IKLink} eliminates the need to manually segment a long or complex trajectory and is beneficial in real-life scenarios that involve end-effector trajectory tracking, such as welding, sweeping, scanning, painting, and inspection.

\vspace{-0.5mm}
\section{Acknowledgement}
We thank Mike Hagenow for fruitful discussions as well as Ben Weinstein for helping with the supplementary video.

\bibliography{root}

\begin{thebibliography}{10}
\providecommand{\url}[1]{#1}
\csname url@samestyle\endcsname
\providecommand{\newblock}{\relax}
\providecommand{\bibinfo}[2]{#2}
\providecommand{\BIBentrySTDinterwordspacing}{\spaceskip=0pt\relax}
\providecommand{\BIBentryALTinterwordstretchfactor}{4}
\providecommand{\BIBentryALTinterwordspacing}{\spaceskip=\fontdimen2\font plus
\BIBentryALTinterwordstretchfactor\fontdimen3\font minus
  \fontdimen4\font\relax}
\providecommand{\BIBforeignlanguage}[2]{{%
\expandafter\ifx\csname l@#1\endcsname\relax
\typeout{** WARNING: IEEEtran.bst: No hyphenation pattern has been}%
\typeout{** loaded for the language `#1'. Using the pattern for}%
\typeout{** the default language instead.}%
\else
\language=\csname l@#1\endcsname
\fi
#2}}
\providecommand{\BIBdecl}{\relax}
\BIBdecl

\bibitem{kang2020torm}
M.~Kang, H.~Shin, D.~Kim, and S.-E. Yoon, ``Torm: Fast and accurate trajectory
  optimization of redundant manipulator given an end-effector path,'' in
  \emph{2020 IEEE/RSJ International Conference on Intelligent Robots and
  Systems (IROS)}.\hskip 1em plus 0.5em minus 0.4em\relax IEEE, 2020, pp.
  9417--9424.

\bibitem{holladay2016distance}
R.~M. Holladay and S.~S. Srinivasa, ``Distance metrics and algorithms for task
  space path optimization,'' in \emph{2016 IEEE/RSJ International Conference on
  Intelligent Robots and Systems (IROS)}.\hskip 1em plus 0.5em minus
  0.4em\relax IEEE, 2016, pp. 5533--5540.

\bibitem{rakita2019stampede}
D.~Rakita, B.~Mutlu, and M.~Gleicher, ``Stampede: A discrete-optimization
  method for solving pathwise-inverse kinematics,'' in \emph{2019 International
  Conference on Robotics and Automation (ICRA)}.\hskip 1em plus 0.5em minus
  0.4em\relax IEEE, 2019, pp. 3507--3513.

\bibitem{yang2022optimal}
T.~Yang, J.~V. Miro, Y.~Wang, and R.~Xiong, ``Optimal task-space tracking with
  minimum manipulator reconfiguration,'' \emph{IEEE Robotics and Automation
  Letters}, vol.~7, no.~2, pp. 5079--5086, 2022.

\bibitem{cefalo2013task}
M.~Cefalo, G.~Oriolo, and M.~Vendittelli, ``Task-constrained motion planning
  with moving obstacles,'' in \emph{2013 IEEE/RSJ International Conference on
  Intelligent Robots and Systems}.\hskip 1em plus 0.5em minus 0.4em\relax IEEE,
  2013, pp. 5758--5763.

\bibitem{siciliano1990kinematic}
B.~Siciliano, ``Kinematic control of redundant robot manipulators: A
  tutorial,'' \emph{Journal of intelligent and robotic systems}, vol.~3, pp.
  201--212, 1990.

\bibitem{praveena2019user}
P.~Praveena, D.~Rakita, B.~Mutlu, and M.~Gleicher, ``User-guided offline
  synthesis of robot arm motion from 6-dof paths,'' in \emph{2019 International
  Conference on Robotics and Automation (ICRA)}.\hskip 1em plus 0.5em minus
  0.4em\relax IEEE, 2019, pp. 8825--8831.

\bibitem{yoon2023learning}
M.~Yoon, M.~Kang, D.~Park, and S.-E. Yoon, ``Learning-based initialization of
  trajectory optimization for path-following problems of redundant
  manipulators,'' in \emph{2023 IEEE International Conference on Robotics and
  Automation (ICRA)}.\hskip 1em plus 0.5em minus 0.4em\relax IEEE, 2023, pp.
  9686--9692.

\bibitem{holladay2019minimizing}
R.~Holladay, O.~Salzman, and S.~Srinivasa, ``Minimizing task-space frechet
  error via efficient incremental graph search,'' \emph{IEEE Robotics and
  Automation Letters}, vol.~4, no.~2, pp. 1999--2006, 2019.

\bibitem{cefalo2013planning}
M.~Cefalo, G.~Oriolo, and M.~Vendittelli, ``Planning safe cyclic motions under
  repetitive task constraints,'' in \emph{2013 IEEE international conference on
  robotics and automation}.\hskip 1em plus 0.5em minus 0.4em\relax IEEE, 2013,
  pp. 3807--3812.

\bibitem{schulman2014motion}
J.~Schulman, Y.~Duan, J.~Ho, A.~Lee, I.~Awwal, H.~Bradlow, J.~Pan, S.~Patil,
  K.~Goldberg, and P.~Abbeel, ``Motion planning with sequential convex
  optimization and convex collision checking,'' \emph{The International Journal
  of Robotics Research}, vol.~33, no.~9, pp. 1251--1270, 2014.

\bibitem{alatartsev2014improving}
S.~Alatartsev and F.~Ortmeier, ``Improving the sequence of robotic tasks with
  freedom of execution,'' in \emph{2014 IEEE/RSJ International Conference on
  Intelligent Robots and Systems}.\hskip 1em plus 0.5em minus 0.4em\relax IEEE,
  2014, pp. 4503--4510.

\bibitem{Descartes}
\BIBentryALTinterwordspacing
ROS-I. (2015) Descartes—a ros-industrial project for performing path-planning
  on under-defined cartesian trajectories. [Online]. Available:
  \url{http://wiki.ros.org/descartes}
\BIBentrySTDinterwordspacing

\bibitem{niyaz2020following}
S.~Niyaz, A.~Kuntz, O.~Salzman, R.~Alterovitz, and S.~Srinivasa, ``Following
  surgical trajectories with concentric tube robots via nearest-neighbor
  graphs,'' in \emph{Proceedings of the 2018 International Symposium on
  Experimental Robotics}.\hskip 1em plus 0.5em minus 0.4em\relax Springer,
  2020, pp. 3--13.

\bibitem{oriolo2002probabilistic}
G.~Oriolo, M.~Ottavi, and M.~Vendittelli, ``Probabilistic motion planning for
  redundant robots along given end-effector paths,'' in \emph{IEEE/RSJ
  International Conference on Intelligent Robots and Systems}, vol.~2.\hskip
  1em plus 0.5em minus 0.4em\relax IEEE, 2002, pp. 1657--1662.

\bibitem{malhan2022generation}
R.~K. Malhan, S.~Thakar, A.~M. Kabir, P.~Rajendran, P.~M. Bhatt, and S.~K.
  Gupta, ``Generation of configuration space trajectories over semi-constrained
  cartesian paths for robotic manipulators,'' \emph{IEEE Transactions on
  Automation Science and Engineering}, vol.~20, no.~1, pp. 193--205, 2022.

\bibitem{yang2020cellular}
T.~Yang, J.~V. Miro, Q.~Lai, Y.~Wang, and R.~Xiong, ``Cellular decomposition
  for nonrepetitive coverage task with minimum discontinuities,''
  \emph{IEEE/ASME Transactions on Mechatronics}, vol.~25, no.~4, pp.
  1698--1708, 2020.

\bibitem{yang2020non}
T.~Yang, J.~V. Miro, Y.~Wang, and R.~Xiong, ``Non-revisiting coverage task with
  minimal discontinuities for non-redundant manipulators,'' in \emph{16th
  Conference on Robotics-Science and Systems}.\hskip 1em plus 0.5em minus
  0.4em\relax MIT PRESS, 2020.

\bibitem{beeson2015trac}
P.~Beeson and B.~Ames, ``Trac-ik: An open-source library for improved solving
  of generic inverse kinematics,'' in \emph{2015 IEEE-RAS 15th International
  Conference on Humanoid Robots (Humanoids)}.\hskip 1em plus 0.5em minus
  0.4em\relax IEEE, 2015, pp. 928--935.

\bibitem{rakita2018relaxedik}
D.~Rakita, B.~Mutlu, and M.~Gleicher, ``Relaxedik: Real-time synthesis of
  accurate and feasible robot arm motion.'' in \emph{Robotics: Science and
  Systems}, vol.~14.\hskip 1em plus 0.5em minus 0.4em\relax Pittsburgh, PA,
  2018, pp. 26--30.

\bibitem{mcinnes2018umap}
L.~McInnes, J.~Healy, and J.~Melville, ``Umap: Uniform manifold approximation
  and projection for dimension reduction,'' \emph{arXiv preprint
  arXiv:1802.03426}, 2018.

\bibitem{ester1996density}
M.~Ester, H.-P. Kriegel, J.~Sander, X.~Xu \emph{et~al.}, ``A density-based
  algorithm for discovering clusters in large spatial databases with noise,''
  in \emph{kdd}, vol.~96, no.~34, 1996, pp. 226--231.

\bibitem{wang2023rangedik}
Y.~Wang, P.~Praveena, D.~Rakita, and M.~Gleicher, ``Rangedik: An
  optimization-based robot motion generation method for ranged-goal tasks,'' in
  \emph{2023 IEEE International Conference on Robotics and Automation
  (ICRA)}.\hskip 1em plus 0.5em minus 0.4em\relax IEEE, 2023, pp. 8090--8096.

\bibitem{kim1995general}
M.-J. Kim, M.-S. Kim, and S.~Y. Shin, ``A general construction scheme for unit
  quaternion curves with simple high order derivatives,'' in \emph{Proceedings
  of the 22nd annual conference on Computer graphics and interactive
  techniques}, 1995, pp. 369--376.

\bibitem{karaman2011sampling}
S.~Karaman and E.~Frazzoli, ``Sampling-based algorithms for optimal motion
  planning,'' \emph{The international journal of robotics research}, vol.~30,
  no.~7, pp. 846--894, 2011.

\bibitem{ames2022ikflow}
B.~Ames, J.~Morgan, and G.~Konidaris, ``Ikflow: Generating diverse inverse
  kinematics solutions,'' \emph{IEEE Robotics and Automation Letters}, vol.~7,
  no.~3, pp. 7177--7184, 2022.

\end{thebibliography}
\bibliographystyle{IEEEtran}

\end{document}